\documentclass[runningheads]{llncs}
\usepackage{amsfonts}
\usepackage{booktabs}
\usepackage{graphicx}

\begin{document}
\title{Neural POS Tagging for the Low-Resource Saraiki}

\author{
Usman Nawaz\inst{1} \and
Muhammad Junaid Iqbal\inst{2} \and
Muhammad Asaf\inst{3} \and
Mariam Munsif Mir\inst{4} \and
Faisal Rehman\inst{5} \and
Tahir Alyas\inst{6}\and
Muhammad Hasnain\inst{7}
}

\authorrunning{U. Nawaz et al.}

\institute{
Independent Researcher, Palermo, Italy\\
\and
University of Rome Tor Vergata, Rome, Italy
\email{Muhammad.junaid.iqbal@uniroma2.it}
\and
DIMES, University of Calabria, Rende, Italy
\and
School of ICT, Griffith University, Nathan, QLD 4111, Australia
\and
University of Mianwali, Mianwali, Pakistan
\and
Department of Computer Science, Lahore Garrison University, Lahore, Pakistan
\and
School of Computer Science, Minhaj University, Lahore, Pakistan
}




\maketitle              
\begin{abstract}

Part-of-speech tagging is a fundamental sequence labeling task and an important component for many downstream natural language processing applications. However, Saraiki, written from right to left in a Perso-Arabic script, remains an under-resourced Indo-Aryan language, despite being spoken as a mother tongue by 28,849,579 people (approximately 28.9 million) according to Pakistan's 7th Population and Housing Census 2023. This paper presents a manually annotated Saraiki Universal POS tagging dataset developed from plain-text. We annotated the text with 17 POS categories. We evaluate three pretrained transformer encoders under the same fine-tuning setup: XLM-RoBERTa base, Urdu RoBERTa small, and multilingual DistilBERT. Experimental results over five random seeds show that XLM-RoBERTa
base achieves the best performance, with 87.29\% $\pm$ 0.21\%
development accuracy and 88.60\% $\pm$ 0.21\% test accuracy. Urdu RoBERTa small and multilingual DistilBERT also provide competitive results, suggesting that multilingual and related-script pretrained encoders are effective for Saraiki UPOS tagging. The dataset, evaluation setup, and baseline results introduced in this work provide a benchmark for future research on Saraiki NLP and low-resource sequence labeling.



\end{abstract}

\section{Introduction}
\label{sec:introduction}

Part-of-speech tagging is a fundamental sequence labeling task that assigns grammatical categories such as noun, adjective, verb, and pronoun to tokens in a sentence. POS tags are useful for many downstream NLP tasks, including machine translation, syntactic parsing, named entity recognition, information extraction and linguistic corpus development \cite{tseng2005morphological}. For low-resource languages, POS-tagged corpora are especially important because they provide a basic layer of linguistic annotation for building and evaluating NLP tools \cite{hedderich2021survey}.

Saraiki, also spelled Siraiki, is an Indo-Aryan language spoken mainly in Pakistan, particularly in Southern Punjab and neighboring regions \cite{alam2024universal}. According to Pakistan's 7th Population and Housing Census 2023, Saraiki is reported as the mother tongue of 28,849,579 people, approximately 28.9 million, representing 12.00\% of Pakistan's population \cite{pbs2023nationalcensus}. It is written from right to left in a Perso-Arabic script and is closely related to Punjabi and Sindhi. Despite this sizeable speaker community and its rich literary and oral tradition, Saraiki remains under-resourced in computational linguistics. Compared with high-resource languages, Saraiki has limited annotated benchmark datasets and fewer publicly available NLP tools. This lack of resources makes it challenging to develop robust systems for automatic language processing \cite{hedderich2021survey}.

Recent work has begun to address this gap. Alam et al. introduced the first Universal Dependencies treebank for Saraiki, containing 587 sentences and 7,597 tokens annotated using the Universal Dependencies framework \cite{alam2024universal}. Their work describes important syntactic and morphological properties of Saraiki, including split ergativity, light verb constructions, serial verb constructions, compounds, and relative clauses. We could not locate a public release of their annotated treebank. Other studies have proposed Saraiki-specific POS tagsets and linguistic annotation schemes. However, publicly available Saraiki UPOS resources and reproducible benchmarks for automatic POS tagging remain limited and there is still a need for additional manually annotated data and reproducible benchmark evaluations for automatic POS tagging.

Saraiki also presents challenges for NLP because of its linguistic and dialectal diversity. Recent work has shown that Saraiki dialects spoken in regions such as Multan and other areas exhibit noticeable lexical variation \cite{akbar2025examining}. Such variation can affect vocabulary coverage, tokenization, and model generalization \cite{hedderich2021survey}. In addition, Saraiki shares script-level similarities with Urdu and Punjabi Shahmukhi, but it has its own grammatical and lexical properties \cite{sajjad2009tagging,ahmed2015cle,tehseen2023neural}. These factors make it important to evaluate NLP models directly on Saraiki data rather than assuming that resources or models developed for related languages will transfer without loss \cite{ruder2019transfer}.

Universal POS tagging provides a useful framework for developing such resources. The Universal POS tagset offers a cross-linguistically consistent set of coarse-grained POS categories \cite{mcdonald2013universal}. It is widely used in the Universal Dependencies project, which provides a common framework for morphological and syntactic annotation across languages \cite{de2021universal,nivre2020universal}. Using UPOS annotation for Saraiki makes the resource compatible with multilingual NLP research and with existing UD-style resources for related Indo-Aryan languages.

In this paper, we address the lack of benchmark resources for Saraiki POS tagging by introducing a manually annotated Universal POS tagging dataset. The corpus is developed from plain-text Saraiki data and annotated with the Universal POS tag inventory \cite{de2021universal,nivre2020universal}. It contains 17 UPOS categories and is divided into fixed train, development, and test splits to support reproducible evaluation.

We further establish transformer-based baseline results for Saraiki UPOS tagging. Specifically, we compare multilingual and related-script pretrained encoders under a consistent experimental setting, including XLM-RoBERTa base, Urdu RoBERTa small, and multilingual DistilBERT.  Our results show that XLM-RoBERTa base performs best, while Urdu RoBERTa small and multilingual DistilBERT provide competitive baselines. These findings suggest that multilingual pretraining and related-script representations are useful for Saraiki sequence labeling.

We summarize the contributions of this paper as follows:

\begin{itemize}
    \item We introduce a manually annotated Saraiki UPOS tagging dataset developed from plain-text Saraiki data, covering 17 Universal POS categories and released splits.

    \item We establish baseline results for Saraiki UPOS tagging by evaluating multilingual and related-script pretrained encoders under a consistent experimental setting.
\end{itemize}

\section{Related Work}
\label{sec:related-work}

POS tagging has been widely studied as a core sequence labeling task in NLP \cite{manning1999foundations,toutanova2003feature,reimers2017reporting}. Early POS tagging systems relied on rule-based methods, statistical models, and feature-based machine learning approaches \cite{brill1992simple,brants2000tnt,toutanova2003feature}. More recent work has shifted toward neural and pretrained transformer-based models, which learn contextualized representations and can be fine-tuned for token-level classification tasks \cite{huang2015bidirectional,lample2016neural,devlin2019bert,nawaz2026shahiemotion}. For low-resource languages, however, the lack of annotated data remains a major limitation \cite{hedderich2021survey}. In such settings, manually annotated corpora and reproducible baseline systems are important for both linguistic research and NLP tool development.

Universal POS tagging provides a common framework for cross-linguistic POS annotation. The Universal POS tagset was introduced to support consistent annotation across languages using a small set of coarse-grained grammatical categories. It is now widely used in the Universal Dependencies framework, which provides annotation guidelines and treebanks for many languages \cite{de2021universal,nivre2020universal}. UD-style annotation has supported multilingual research in parsing, morphology, and sequence labeling. However, coverage remains uneven, and many low-resource languages still lack sufficiently large annotated corpora \cite{hedderich2021survey,nawaz2026towards}.

Several Universal Dependencies resources have been developed for Indo-Aryan languages. Existing UD treebanks include resources for Hindi, Urdu, Marathi, and Punjabi \cite{tandon2016conversion,bhat2012dependency,ehsan2020dependency,ravishankar2017universal,arora2022universal}. These resources show the usefulness of UD annotation for South Asian languages, but they also highlight the uneven availability of linguistic resources across the Indo-Aryan family. Major languages such as Hindi and Urdu have received more attention than smaller or regionally concentrated languages. As a result, many Indo-Aryan languages remain under-represented in computational resources and benchmark evaluations.

Earlier Saraiki NLP work has also addressed POS tagset design and linguistic annotation proposed a POS tagset for morphological, syntactic, and lexical annotation of Saraiki \cite{saleemi2021basic}. Malik et al. \cite{malik2023hierarchical} presented a hierarchical POS tagset for Saraiki with a richer set of language-specific categories. These studies are valuable because they document Saraiki-specific grammatical categories and provide foundations for linguistic annotation. However, language-specific tagsets may not be directly comparable across languages. In contrast, UPOS annotation supports multilingual comparison and compatibility with Universal Dependencies-style resources \cite{de2021universal}.

Saraiki also presents challenges because of dialectal and lexical variation spoken in different regions of Pakistan \cite{akbar2025examining}. Their findings show that lexical choices vary across regional boundaries. Such variation is important for NLP because it can affect vocabulary coverage, tokenization, and model generalization. For POS tagging, dialectal variation may introduce different word forms or region-specific lexical items that are not equally represented in training data.

Transformer-based models have become the dominant approach for many sequence labeling tasks \cite{vaswani2017attention,devlin2019bert}. XLM-RoBERTa is a multilingual transformer trained on an extensive amount of data and has achieved strong results on many cross-lingual tasks \cite{conneau2020unsupervised}. DistilBERT provides a smaller and faster distilled transformer model while retaining much of the performance of the original transformer architecture \cite{sanh2019distilbert}. For Saraiki, related-script models are also relevant because Saraiki is written in Perso-Arabic script and shares script-level similarities with Urdu and Punjabi Shahmukhi \cite{sajjad2009tagging,ahmed2015cle,tehseen2023neural}. Therefore, Urdu-focused pretrained models may provide useful representations for Saraiki text.

Overall, previous work has provided important foundations for Saraiki linguistic annotation, including POS tagsets and a small Universal Dependencies treebank. However, benchmark-style evaluation of pretrained transformer encoders for Saraiki UPOS tagging remains limited. This paper addresses this gap by manually annotating Saraiki plain text with Universal POS tags and evaluating three pretrained transformer models under a fixed train--development--test setup.

\section{Dataset Construction}
\label{sec:dataset}

\subsection{Data Source}
\label{subsec:data-source}

The source data contains Saraiki text written in the Perso-Arabic script \cite{kaleem2025saraiki}. Since the original dataset was plain text, it did not provide token boundaries, POS labels, morphological features, or dependency relations. We therefore manually processed and annotated the data for Universal POS tagging. The goal was to create a resource suitable for training and evaluating automatic Saraiki POS taggers.

Saraiki is written from right to left and shares script-level properties with Urdu and Punjabi Shahmukhi. However, it also has its own lexical, morphological, and syntactic characteristics. For this reason, we kept the text in its original Saraiki script and did not transliterate it into Roman script or any other writing system. This allows the resulting dataset to reflect the form in which Saraiki is commonly written and processed.

\subsection{Preprocessing}
\label{subsec:preprocessing}

Before annotation, we cleaned the raw text by removing empty lines, unusable rows, and duplicate or malformed entries where necessary. We also normalized whitespace while preserving the original script. Since the aim of this work is to evaluate Saraiki POS tagging in its native writing system, no translation or transliteration was applied.

The cleaned sentences were then manually segmented into tokens. Each sentence was converted into CoNLL-U style format, where each token appears on a separate line with its corresponding annotation fields \cite{nivre2020universal}. Since the present work focuses on Universal POS tagging, we annotate the UPOS column and leave the XPOS, morphological-feature, and dependency fields unspecified \cite{de2021universal}.

\subsection{UPOS Label Inventory}
\label{subsec:upos-labels}

We annotate the dataset using the Universal POS tag inventory \cite{nivre2020universal}. The UPOS tagset provides a small set of coarse-grained grammatical categories that can be applied consistently across languages. This makes the dataset compatible with Universal Dependencies-style annotation and allows comparison with POS tagging resources for other languages.

The dataset contains 17 UPOS categories:

\begin{center}
\small
\begin{tabular}{llll}
\hline
ADJ & ADP & ADV & AUX \\
CCONJ & DET & INTJ & NOUN \\
NUM & PART & PRON & PROPN \\
PUNCT & SCONJ & SYM & VERB \\
X & & & \\
\hline
\end{tabular}
\end{center}

Each token is assigned exactly one UPOS label. Tokens that function as common nouns are annotated as \texttt{NOUN}, while proper names are annotated as \texttt{PROPN}. Main lexical verbs are annotated as \texttt{VERB}, and auxiliary elements are annotated as \texttt{AUX}. Adpositions and postpositions are annotated as \texttt{ADP}. Punctuation marks are annotated as \texttt{PUNCT}. The label \texttt{X} is used for tokens that cannot be assigned to one of the regular grammatical categories.

\subsection{Manual Annotation}
\label{subsec:manual-annotation}

The annotation was carried out manually using the Universal POS
tag inventory. One annotator, who holds an MPhil in Linguistics and
has experience in linguistic annotation, assigned a UPOS label to
each token according to its grammatical function in context. Two
PhD students independently reviewed the entire annotated dataset
for correctness and made manual corrections where needed. Difficult
cases were considered in sentence context, since the same surface
form may belong to different grammatical categories depending on
usage.

The annotation process focused on producing a consistent token-level POS resource. Since there is no widely adopted large-scale Saraiki POS benchmark, we followed the Universal POS categories rather than a language-specific tagset. This choice supports cross-linguistic comparability and makes the dataset easier to use with existing multilingual NLP models and Universal Dependencies-oriented tools.

Some annotation decisions are especially relevant for Saraiki. Verbal elements were annotated according to their function as main verbs or auxiliaries. Nouns and proper nouns were distinguished based on whether the token referred to a general entity or a named entity. Pronouns, determiners, and relative forms were annotated according to their syntactic function in the sentence. These decisions are important because Saraiki shares several grammatical patterns with related Indo-Aryan languages, but also has language-specific variation.

\subsection{Data Splits}
\label{subsec:data-splits}

After annotation, the dataset was divided into fixed train, development, and test splits. The training split is used for model fine-tuning, the development split is used for best checkpoint selection, and the test split is reserved for final evaluation. Keeping fixed splits allows future work to compare results under the same evaluation setting. Table~\ref{tab:dataset-statistics} shows the distribution of sentences and tokens across the dataset splits.

\begin{table}[!h]
\centering
\small
\caption{Dataset statistics for the training, development, and test splits.}
\label{tab:dataset-statistics}
\begin{tabular}{lrr}
\toprule
\textbf{Dataset Split} & \textbf{Sentences} & \textbf{Tokens} \\
\midrule
Training     & 1,666 & 28,377 \\
Development  & 208   & 3,636  \\
Test         & 209   & 3,439  \\
\midrule
\textbf{Total} & \textbf{2,083} & \textbf{35,452} \\
\bottomrule
\end{tabular}
\end{table}

\begin{table}[htbp]
\centering
\small
\setlength{\tabcolsep}{4pt}
\caption{Number of tokens per UPOS tag in each data split.}
\label{tab:upos-counts}
\begin{tabular}{lrrrr}
\toprule
\textbf{Tag} & \textbf{Train} & \textbf{Dev} & \textbf{Test} & \textbf{Total} \\
\midrule
ADJ   & 1,430 & 205 & 153 & 1,788 \\
ADP   & 4,235 & 544 & 526 & 5,305 \\
ADV   & 267 & 44 & 32 & 343 \\
AUX   & 1,746 & 208 & 204 & 2,158 \\
CCONJ & 856 & 127 & 129 & 1,112 \\
DET   & 73 & 12 & 8 & 93 \\
INTJ  & 1 & 0 & 0 & 1 \\
NOUN  & 8,915 & 1,241 & 1,050 & 11,206 \\
NUM   & 487 & 69 & 44 & 600 \\
PART  & 511 & 78 & 65 & 654 \\
PRON  & 1,348 & 189 & 174 & 1,711 \\
PROPN & 3,651 & 312 & 495 & 4,458 \\
PUNCT & 2,479 & 284 & 258 & 3,021 \\
SCONJ & 602 & 68 & 73 & 743 \\
SYM   & 16 & 0 & 0 & 16 \\
VERB  & 1,749 & 255 & 228 & 2,232 \\
X     & 11 & 0 & 0 & 11 \\
\midrule
\textbf{Total} & \textbf{28,377} & \textbf{3,636} &
\textbf{3,439} & \textbf{35,452} \\
\bottomrule
\end{tabular}
\end{table}

\section{Methodology}
\label{sec:methodology}


We formulate Saraiki Universal POS tagging as a token-level sequence labeling task. Given an input sentence
\begin{equation}
S = (z_1, z_2, \dots, z_n),
\end{equation}
where $z_i$ indicates the $i$-th token in the sequence, the goal is to assign a UPOS label to each token $z_i$.
\begin{equation}
y_i \in \mathcal{Y}
\end{equation}
Here, $\mathcal{Y}$ is the set of 17 Universal POS categories used in the annotated corpus. Each token receives exactly one UPOS label.

\subsection{Model Architecture}
\label{subsec:model-architecture}

We use pretrained transformer encoders for token classification \cite{vaswani2017attention,devlin2019bert}. Given a tokenized input sentence, the encoder produces contextualized representations for the input tokens. These representations are passed through a dropout layer and a linear classification head to produce logits over the UPOS label set.

For an input token representation $\mathbf{h}_i$, the model computes:
\begin{equation}
\mathbf{o}_i = \mathbf{W}\mathbf{h}_i + \mathbf{b},
\end{equation}
where $\mathbf{o}_i \in \mathbb{R}^{|\mathcal{Y}|}$ is the output logit vector for token $i$, $\mathbf{W}$ and $\mathbf{b}$ are the parameters of the classification head, and $|\mathcal{Y}| = 17$. The predicted UPOS label is obtained by selecting the label with the highest logit:
\begin{equation}
\hat{y}_i = \arg\max_{k \in \mathcal{Y}} o_{i,k}.
\end{equation}

The same architecture is used for all pretrained encoders: a transformer encoder followed by dropout and a linear UPOS classification head. This ensures that the comparison focuses on the pretrained encoder rather than differences in the downstream classifier. All encoder parameters and classification head parameters are fine-tuned during training.

We train all models using standard cross-entropy loss computed over the labeled token positions. Let $\mathcal{I}$ be the set of token positions included in loss computation. The loss is computed as:
\begin{equation}
\mathcal{L}_{\mathrm{CE}} =
-\frac{1}{|\mathcal{I}|}
\sum_{i \in \mathcal{I}}
\log
\frac{\exp(o_{i,y_i})}
{\sum_{k=1}^{|\mathcal{Y}|} \exp(o_{i,k})}.
\end{equation}

\section{Experimental Setup}
\label{sec:experimental-setup}

We evaluate Saraiki UPOS tagging using the fixed splits described in Section~\ref{sec:dataset} and repeat each experiment using five random seeds (42--46) and report the mean and sample standard deviation of accuracy.

\subsection{Models}
\label{subsec:models}

We compare three pretrained transformer encoders under the same
token classification setup.\footnote{Pretrained checkpoints:
XLM-RoBERTa base,
\url{https://huggingface.co/FacebookAI/xlm-roberta-base},
Urdu RoBERTa small,
\url{https://huggingface.co/urduhack/roberta-urdu-small},
multilingual DistilBERT,
\url{https://huggingface.co/distilbert/distilbert-base-multilingual-cased}.} The selected models include multilingual and related-script encoders. XLM-RoBERTa base is included as a strong multilingual model trained on large-scale multilingual data \cite{conneau2020unsupervised}. Urdu RoBERTa small is included because Saraiki and Urdu are both written in Perso-Arabic script and may share useful script-level representations \cite{sajjad2009tagging,ahmed2015cle}. Multilingual DistilBERT is included as a faster and lighter multilingual baseline \cite{sanh2019distilbert}.

All models are fine-tuned using AdamW optimization and a linear learning rate schedule with warmup \cite{loshchilov2019decoupled}. We use a batch size of 16, a maximum sequence length of 256, and train for up to 10 epochs. Early stopping is applied when development accuracy does not improve for four consecutive epochs. The dropout probability is set to 0.10.

\begin{table}[!h]
\centering
\small
\setlength{\tabcolsep}{6pt}
\caption{Hyperparameter settings for fine-tuning.}
\label{tab:hyperparameters}
\begin{tabular}{ll}
\hline
\textbf{Hyperparameter} & \textbf{Value} \\
\hline
Maximum sequence length & 256 \\
Batch size & 16 \\
Epochs & 10 \\
Learning rate & $2 \times 10^{-5}$ \\
Weight decay & 0.01 \\
Dropout & 0.10 \\
Optimizer & AdamW \\
Training objective & Cross-entropy \\
\hline
\end{tabular}
\end{table}

We implemented all experiments in PyTorch and used the Hugging Face Transformers library \cite{paszke2019pytorch,wolf2020transformers}. The models are fine-tuned, meaning that both the pretrained encoder and the UPOS classification head are updated during training.

\subsection{Evaluation Metrics}
\label{subsec:evaluation-metric}

We use UPOS accuracy for the main model comparison. For a detailed
analysis of the XLM-RoBERTa checkpoint selected by development
accuracy, we also report per-class precision, recall, F1-score,
support, and a confusion matrix.

Let $N$ be the number of evaluated tokens, $y_i$ the gold UPOS tag
of token $i$, and $\hat{y}_i$ its predicted tag. Accuracy is the
proportion of correctly tagged tokens:
\begin{equation}
\mathrm{Accuracy} =
\frac{1}{N}\sum_{i=1}^{N}\mathbf{1}\{\hat{y}_i=y_i\},
\end{equation}
where $\mathbf{1}\{\cdot\}$ equals 1 when its condition is true and
0 otherwise.

For each UPOS tag $c$, let $\mathrm{TP}_c$ be the number of tokens
correctly predicted as $c$, $\mathrm{FP}_c$ the number incorrectly
predicted as $c$, and $\mathrm{FN}_c$ the number of gold $c$ tokens
predicted as another tag. Per-class precision, recall, and F1-score
are defined as:
\begin{equation}
P_c=\frac{\mathrm{TP}_c}{\mathrm{TP}_c+\mathrm{FP}_c},
\qquad
R_c=\frac{\mathrm{TP}_c}{\mathrm{TP}_c+\mathrm{FN}_c},
\qquad
F1_c=\frac{2P_cR_c}{P_c+R_c}.
\end{equation}
The support of tag $c$ is its number of gold test tokens,
$n_c=\mathrm{TP}_c+\mathrm{FN}_c$.

We report macro averages over the full set $\mathcal{Y}$ of 17 UPOS
tags. For $M_c\in\{P_c,R_c,F1_c\}$, the macro and support-weighted
averages are:
\begin{equation}
M_{\mathrm{macro}}=
\frac{1}{|\mathcal{Y}|}\sum_{c\in\mathcal{Y}}M_c,
\qquad
M_{\mathrm{weighted}}=
\frac{1}{N}\sum_{c\in\mathcal{Y}}n_cM_c,
\quad |\mathcal{Y}|=17.
\end{equation}
Following the evaluation implementation, a metric with a zero
denominator is assigned zero. INTJ, SYM, and X have no gold test
examples; their zero scores contribute to the 17-tag macro average
but do not constitute measured performance on those tags.

\section{Results and Analysis}
\label{sec:results}

Table~\ref{tab:main-results} presents the mean development and test
accuracy of the three evaluated transformer models over five random
seeds. For each run, the checkpoint with the highest development
accuracy was selected and evaluated on the held-out test set.

\begin{table}[!h]
\centering
\small
\caption{UPOS accuracy (\%) over five random seeds, reported as mean
$\pm$ sample standard deviation.}
\label{tab:main-results}
\begin{tabular}{lcc}
\toprule
\textbf{Model} & \textbf{Dev} & \textbf{Test} \\
\midrule
XLM-RoBERTa base & \textbf{87.29 $\pm$ 0.21}
                  & \textbf{88.60 $\pm$ 0.21} \\
Urdu RoBERTa small & 85.11 $\pm$ 0.14
                    & 86.85 $\pm$ 0.21 \\
Multilingual DistilBERT & 84.44 $\pm$ 0.32
                         & 86.03 $\pm$ 0.30 \\
\bottomrule
\end{tabular}
\end{table}

\begin{table}[h]
\centering
\small
\setlength{\tabcolsep}{4pt}
\caption{Per-class test results for XLM-RoBERTa (seed 46).
Support is the number of gold test tokens.}
\label{tab:per-class-results}
\begin{tabular}{lrrrr}
\toprule
\textbf{UPOS} & \textbf{Precision} & \textbf{Recall}
& \textbf{F1} & \textbf{Support} \\
\midrule
ADJ   & 0.6826 & 0.7451 & 0.7125 & 153 \\
ADP   & 0.9616 & 0.9525 & 0.9570 & 526 \\
ADV   & 0.7083 & 0.5312 & 0.6071 & 32 \\
AUX   & 0.8883 & 0.8971 & 0.8927 & 204 \\
CCONJ & 0.8603 & 0.9070 & 0.8830 & 129 \\
DET   & 0.4615 & 0.7500 & 0.5714 & 8 \\
INTJ  & 0.0000 & 0.0000 & 0.0000 & 0 \\
NOUN  & 0.8914 & 0.8676 & 0.8793 & 1,050 \\
NUM   & 0.8478 & 0.8864 & 0.8667 & 44 \\
PART  & 0.9655 & 0.8615 & 0.9106 & 65 \\
PRON  & 0.8922 & 0.8563 & 0.8739 & 174 \\
PROPN & 0.8854 & 0.9051 & 0.8951 & 495 \\
PUNCT & 1.0000 & 1.0000 & 1.0000 & 258 \\
SCONJ & 0.8118 & 0.9452 & 0.8734 & 73 \\
SYM   & 0.0000 & 0.0000 & 0.0000 & 0 \\
VERB  & 0.7739 & 0.7807 & 0.7773 & 228 \\
X     & 0.0000 & 0.0000 & 0.0000 & 0 \\
\midrule
Macro avg.\ (17 tags) & 0.6842 & 0.6992 & 0.6882 & 3,439 \\
Weighted avg. & 0.8875 & 0.8857 & 0.8861 & 3,439 \\
\bottomrule
\end{tabular}
\end{table}

Table~\ref{tab:per-class-results} shows that INTJ, SYM, and X have no gold examples in the test set. Their
undefined per-class scores were set to zero when computing the
17-tag macro averages; the zeros do not represent measured
performance on those tags.

\begin{table}[htbp]
\centering
\caption{Test-set UPOS confusion matrix for XLM-RoBERTa (seed 46).
Rows are gold tags; columns are predicted tags.}
\label{tab:confusion-matrix}
\resizebox{\textwidth}{!}{%
\begin{tabular}{l*{17}{r}}
\toprule
Gold & AUX & NUM & NOUN & CCONJ & VERB & ADP & PUNCT & ADJ
& PART & PROPN & PRON & SCONJ & ADV & DET & SYM & X & INTJ \\
\midrule
AUX   & 183&0&1&0&19&0&0&0&1&0&0&0&0&0&0&0&0 \\
NUM   & 0&39&3&0&0&0&0&0&0&0&0&0&2&0&0&0&0 \\
NOUN  & 4&4&911&1&23&7&0&30&0&53&11&4&1&1&0&0&0 \\
CCONJ & 0&0&0&117&0&10&0&0&0&0&0&2&0&0&0&0&0 \\
VERB  & 13&0&26&0&178&0&0&5&1&1&3&0&0&1&0&0&0 \\
ADP   & 0&1&5&17&1&501&0&1&0&0&0&0&0&0&0&0&0 \\
PUNCT & 0&0&0&0&0&0&258&0&0&0&0&0&0&0&0&0&0 \\
ADJ   & 0&2&26&0&3&0&0&114&0&4&1&1&0&2&0&0&0 \\
PART  & 1&0&3&0&1&1&0&1&56&0&0&1&1&0&0&0&0 \\
PROPN & 1&0&31&0&3&0&0&12&0&448&0&0&0&0&0&0&0 \\
PRON  & 1&0&9&0&0&0&0&3&0&0&149&7&2&3&0&0&0 \\
SCONJ & 1&0&2&1&0&0&0&0&0&0&0&69&0&0&0&0&0 \\
ADV   & 2&0&5&0&2&2&0&0&0&0&3&1&17&0&0&0&0 \\
DET   & 0&0&0&0&0&0&0&1&0&0&0&0&1&6&0&0&0 \\
SYM   & 0&0&0&0&0&0&0&0&0&0&0&0&0&0&0&0&0 \\
X     & 0&0&0&0&0&0&0&0&0&0&0&0&0&0&0&0&0 \\
INTJ  & 0&0&0&0&0&0&0&0&0&0&0&0&0&0&0&0&0 \\
\bottomrule
\end{tabular}%
}
\end{table}

Table~\ref{tab:confusion-matrix} presents the test-set confusion
matrix for the XLM-RoBERTa checkpoint selected by development
accuracy. Rows represent gold UPOS tags, and columns represent
predicted tags. The model correctly labels 3,046 of 3,439 tokens.
The most frequent errors are NOUN predicted as PROPN (53 tokens),
PROPN predicted as NOUN (31), and NOUN predicted as ADJ (30).
The INTJ, SYM, and X rows contain no examples because these tags
are absent from the test set.

The results show that the choice of pretrained encoder has a clear
effect on Saraiki UPOS tagging performance. XLM-RoBERTa base achieves
the strongest mean result, with 87.29\% development accuracy and
88.60\% test accuracy. Urdu RoBERTa small obtains the second-best
mean test result at 86.85\%, followed by multilingual DistilBERT
at 86.03\%.

The ranking of the models is consistent across the development and test sets. XLM-RoBERTa base performs best on both splits, followed by Urdu RoBERTa small and multilingual DistilBERT. This consistency suggests that the observed differences are not specific to a single split.

XLM-RoBERTa base improves over multilingual DistilBERT by
2.57 percentage points on the test set and over Urdu RoBERTa
small by 1.75 percentage points. These results indicate that large-scale multilingual pretraining provides a measurable advantage for Saraiki UPOS tagging. At the same time, Urdu RoBERTa small remains competitive despite being smaller than XLM-RoBERTa base, suggesting that related-script pretraining can also be useful for Saraiki text.

Multilingual DistilBERT obtains the lowest accuracy among the three models, but it still provides a competitive lightweight baseline. Its result shows that smaller multilingual encoders can be useful when computational efficiency is important, although they may not match the accuracy of larger multilingual models.

Overall, the results suggest that multilingual and related-script pretrained encoders are suitable for Saraiki UPOS tagging. XLM-RoBERTa base is the strongest model under the current setup, while Urdu RoBERTa small and multilingual DistilBERT provide useful alternative baselines.

\section{Discussion}
\label{sec:discussion}

The results demonstrate that pretrained transformer encoders can be effective for Saraiki UPOS tagging, but the choice of encoder remains important. The strongest performance is achieved by XLM-RoBERTa base, which suggests that broad multilingual pretraining and wider subword coverage are beneficial for low-resource token classification.

The performance of Urdu RoBERTa small is also notable. Although Urdu and Saraiki are different languages, both are written in Perso-Arabic script and share some orthographic patterns. This may help an Urdu-focused model represent Saraiki tokens more effectively than expected. However, Urdu RoBERTa small still remains below XLM-RoBERTa base, indicating that related-script pretraining is helpful but does not fully replace the benefits of broader multilingual pretraining.

These findings also highlight the value of building annotated resources for Saraiki. Existing Saraiki NLP resources remain limited, and manually annotated UPOS data can support the development and evaluation of automatic taggers. The dataset can also serve as a foundation for future work on morphological analysis, dependency parsing, and other downstream NLP tasks.

At the same time, the current work focuses only on coarse-grained UPOS tagging. Saraiki has rich morphology and several language-specific grammatical phenomena, including light verb constructions, serial verb constructions, split ergativity, and dialectal variation. Future work should therefore extend the corpus with morphological features, language-specific POS tags, and dependency relations.

Overall, the dataset and baseline results provide a starting point for more systematic research on Saraiki NLP and low-resource sequence labeling.

\section{Conclusion}
\label{sec:conclusion}

This paper introduced a manually annotated Saraiki Universal POS tagging dataset developed from plain-text Saraiki data. The corpus was annotated with the Universal POS tag inventory and converted into CoNLL-U format, providing fixed train, development, and test splits for reproducible evaluation.

We established transformer-based baselines for Saraiki UPOS tagging by fine-tuning XLM-RoBERTa base, Urdu RoBERTa small, and multilingual DistilBERT under a consistent token classification setup. The results show that multilingual and related-script pretrained encoders are effective for Saraiki sequence labeling, with XLM-RoBERTa base providing the strongest overall performance among the evaluated models.

Overall, this work provides a benchmark resource and baseline results for Saraiki POS tagging. The dataset can support future research on low-resource NLP, Universal Dependencies-oriented annotation, and downstream Saraiki language processing. Future work will expand the corpus with additional data, morphological features, and dependency relations, and will explore further modeling approaches such as CRF decoding, parameter-efficient fine-tuning, and data augmentation.

\section*{Limitations}
\label{sec:limitations}

This study has several limitations. First, the dataset is limited in size compared with POS-tagged corpora available for high-resource languages. Although the manually annotated corpus provides a useful benchmark for Saraiki UPOS tagging, larger datasets would likely improve model performance and support more reliable evaluation across different genres and dialects.

Second, the present work focuses only on Universal POS tags. UPOS categories are useful for cross-linguistic comparison, but they do not capture the full range of Saraiki-specific morphological and syntactic information. Saraiki has rich verbal morphology, case marking, light verb constructions, serial verb constructions, and other language-specific phenomena. Future versions of the corpus should include morphological features, language-specific POS tags, and dependency relations.

Third, INTJ, SYM, and X have no examples in the test set, so their
test performance cannot be assessed. The per-class analysis is
reported for one checkpoint selected using development accuracy;
the main accuracy comparison is reported over five random seeds.

Fourth, the model comparison is limited to three pretrained transformer encoders. These models provide useful multilingual and related-script baselines, but additional models should be evaluated in future work. This includes larger multilingual encoders, regional language models, Urdu and Punjabi-focused models, and parameter-efficient fine-tuning methods.

Finally, the annotation process may involve difficult linguistic decisions, especially for ambiguous tokens whose POS category depends on context. Since Saraiki also has dialectal and lexical variation, some forms may be used differently across regions. Although two experts reviewed the full corpus, the same sentences
were not independently annotated from scratch by multiple annotators,
and inter-annotator agreement was not measured. Future work should
document more detailed annotation guidelines and use independent
annotation of a representative subset to measure agreement.

\section*{Data Availability}


The annotated Saraiki UPOS corpus and its fixed training,
development, and test splits are available at
\url{https://github.com/ancientintelligencelab/saraikiupos}.

\section*{Ethics Statement}

This work uses Saraiki plain-text data and manually annotates it for Universal POS tagging. The dataset is intended for research on low-resource NLP and language technology development. The corpus may reflect biases, dialectal preferences, or domain-specific patterns present in the original text source. Models trained on this dataset should therefore be used with care, especially outside the domain represented by the data. The resource and models are not intended for high-stakes decision-making.

%
%
%
%





\bibliographystyle{splncs04}
\bibliography{custom}

\end{document}